\documentclass[letterpaper, 10 pt, conference]{ieeeconf}
\IEEEoverridecommandlockouts    
\usepackage[textwidth=4cm,disable]{todonotes}

\usepackage{graphics}           
\usepackage{times}              
\usepackage{amsmath}            
\usepackage{amssymb}            
\usepackage{graphicx}
\usepackage{algorithm}
\usepackage[noend]{algpseudocode}
\usepackage{booktabs}
\usepackage{color}

\definecolor{instructioncolor}{rgb}{.5,.5,.5}

\usepackage[font=small]{caption}

\def\secref#1{Sec.~\ref{#1}}
\def\figref#1{Fig.~\ref{#1}}

\def\eqref#1{Eq.~(\ref{#1})}

\makeatletter
\usepackage{xspace}
\DeclareRobustCommand\onedot{\futurelet\@let@token\@onedot}
\def\@onedot{\ifx\@let@token.\else.\null\fi\xspace}

\makeatother

\usepackage{array}
\newcolumntype{L}[1]{>{\raggedright\let\newline\\\arraybackslash\hspace{0pt}}m{#1}}
\newcolumntype{C}[1]{>{\centering\let\newline\\\arraybackslash\hspace{0pt}}m{#1}}
\newcolumntype{R}[1]{>{\raggedleft\let\newline\\\arraybackslash\hspace{0pt}}m{#1}}

\usepackage{subfigure}
\usepackage{siunitx}

\title{\LARGE \bf Human-Inspired Long-Term Indoor Localization in Human-Oriented Environment}

\author{Nicky Zimmerman\dag  \and Matteo Sodano\ddag
\thanks{\dag IDSIA, Universita della Svizzera italiana, Switzerland
  }%
  \thanks{\ddag University of Bonn, Germany
  }
} 

\begin{document}   
\maketitle
\thispagestyle{empty}
\pagestyle{empty}

\begin{abstract}

Lifelong localization is crucial for enabling the autonomy of service robots. In this paper, we present an overview of our past research on long-term localization and mapping, exploiting geometric priors such as floor plans and integrating textual and semantic information. Our approach was validated on challenging sequences spanning over many months, and we released open source implementations.
  %
 
    
  
 
\end{abstract}

\section{Introduction} \label{sec:intro}

Localization has been long-studied in robotics~\cite{cadena2016tro,thrun2005probrobbook,zafari2019cst}. Supported by high-resolution sensors, GNSS-denied robot localization research initially focused on precise, detailed geometric maps~\cite{bylow2013rss, grisetti2007tro, moravec1989sdsr, newcombe2011ismar}. However, humans do not require highly accurate geometric information to navigate~\cite{mendez2018icra, yi2009iros}, but instead, extract semantic information from objects around them. 
Inspired by how humans navigate, we can exploit insights from human navigation to improve long-term localization, which enables robots to navigate in the same environment over extended periods, spanning several months or even years.
In this work, we summarize our past contributions to robust long-term localization and mapping, exploiting long-lasting geometric, textual and semantic cues~\cite{zimmerman2023ral, zimmerman2024icra, zimmerman2023iros, zimmerman2022iros}. 

Localization research has advanced beyond purely geometric research, and recently many works tackle the task of semantic localization and mapping~\cite{rosinol2020icra, tao2024ral, wada2020cvpr}. However, they focus on getting a detailed reconstruction of the current environment, rather than long-term localization. Human-occupied environments tend to be dynamic, having both fast-moving agents and structural and quasi-static changes such as opening/closing doors and furniture changes. 
When localizing with a known map, similar to how humans revisit places, a critical component is which elements are contained in the map. Regardless of the localization strategy, a map containing unstable features may lead to localization failures. 
To support long-term localization, map elements must be selected rather than including all current observations. These elements can either be inferred with common sense or learned statistically from observing an environment from a period of time~\cite{zimmerman2023ral}. Textual cues are another element that can effectively support long-term localization. Textual information such as room numbers can serve as a unique identifier of places and is often included in floor plans. However, they are often neglected in mapping approaches. Works on text-guided localization are also scarce~\cite{cui2021iros, radwan2016icra}. Thus, we proposed a framework for integrating textual information in maps and exploiting it during localization \cite{zimmerman2022iros}.

A crucial aspect of long-term localization is the level of accuracy required. In fact, there is a trade-off between accuracy and robustness, and each task requires a different blend of the two. For example, for planning and navigating along the path of hundreds of meters, robustness (i.e., avoiding jumps in the trajectory) is more important, while high accuracy is only required in specific end-points (i.e. pickup/delivery points). 
When the requirement of exceptional accuracy is lifted, we can consider less precise maps, such as hand-drawn maps~\cite{li2020iros}, visitor map and floor plans annotated by end users with objects~\cite{loo2024arxiv, zimmerman2023ral}. This relaxation of requirements can even remove the need for mapping, and allow end-users to independently edit floor-plan based maps. 

In our previous works, we advocated for fusing geometric, semantic, and textual information for long-term localization in human-oriented environments. In this paper, we provide a summary of our contributions and insights.

\begin{figure}[t]
  \centering
  \includegraphics[trim={0 4.9cm 0 0},clip, width=0.85\linewidth]{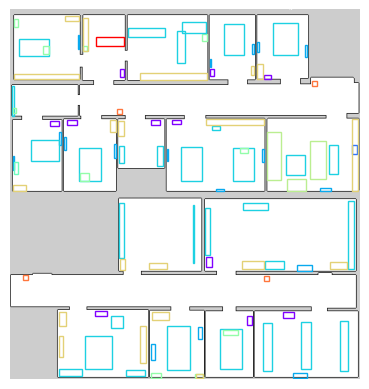}
  \caption{A 2D abstract semantic map enriching a floor plan with semantic information, used for long term localization. Different box colors indicate different object classes.}
  \label{fig:semmaps} 
\end{figure} 

\section{Approach} \label{sec:approach}
Our goal is to globally localize in an indoor environment, exploiting semantic and textual cues, and prior geometric information, when available. We utilize semantically-enriched floor plans that capture long-lasting features, and localize by integrating semantic and textual information in a Monte Carlo localization~(MCL) framework~\cite{dellaert1999icra}.

\subsection{Abstract Semantic map}
We propose the concept of abstract semantic map, a map representation that captures the most long-lasting structures of an environment. 
This map includes a floor plan (or an occupancy grid map), which is overlaid by layers of semantic and textual information. Objects with semantic significance are represented at high-level by their semantic class and 2D rectangles overlying the occupancy grid. Textual information is encoded through the text likelihood maps~(\secref{sec:textlikelihood}). The floor plan is segmented into rooms, where each room contains a list of semantic objects, a room category representing a higher level of semantic understanding and a name corresponding to a text sign.
While we offer a mapping approach that can construct a 3D metric-semantic map from RGB images~\cite{zimmerman2023iros}, it is also possible to edit the abstract semantic map manually, using our GUI application, MAPhisto\footnote{https://github.com/FullMetalNicky/Maphisto}. 


To determine which semantic classes hold significance for the purpose of long-term localization, we estimated how transient instances of these classes were. As part of our work ~\cite{zimmerman2023ral}, we collected data for several months in a dynamic environment of a university lab, including office rooms, corridors and a kitchen. Employing an object detection model on images with known poses, we were able to detect when objects have been moved, and assign a stability score to a semantic class based on how often instances of that classes moved around.


\subsection{Text Likelihood Maps} \label{sec:textlikelihood}


\begin{figure}
  \centering
   \subfigure[]{\includegraphics[trim={0 0 1cm 0}, clip, width=0.8\columnwidth]{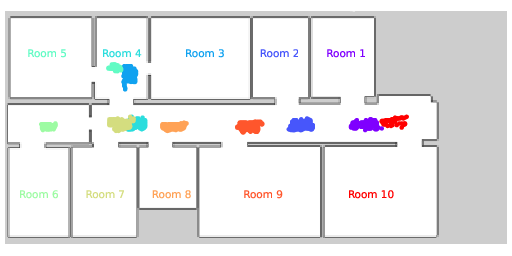}}
  \subfigure[]{\includegraphics[width=0.4\columnwidth]{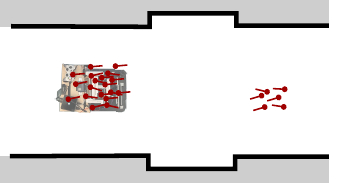}}
  \subfigure[]{\includegraphics[width=0.4\columnwidth]{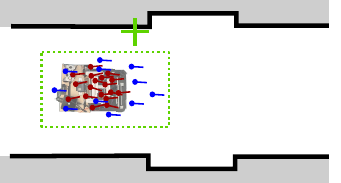}}

  \caption{(a) The text likelihood maps (b) Particle distribution prior to text spotting, with multi-modal distribution (c) Particle injection based on the text likelihood map after a textual cue was detected.}
  \label{fig:textmaps}
  \vspace{-5pt}
\end{figure}

Textual information is integrated into our semantic maps through likelihood functions, describing where the robot might be when detecting a specific textual cue. In our work~\cite{zimmerman2022iros}, we adopt a text spotting pipeline based on the work of Liao et al.~\cite{liao2020aaai} and Shi et al.~\cite{shi2016tpami}, who proposed a detection and recognition models for text, respectively.
We decide to focus on room numbers as a textual localization cue, which can be either assigned manually to the abstract semantic map, or extracted from a floor plan. We record posed images and apply the text spotting, matching each detection to our list of rooms. 
For each room number, we construct a 2D histogram of locations where the text was successfully detected. These sampled locations provide a sparse representation of the likelihood of detecting a specific textual tag, which we refer to as text likelihood maps (\figref{fig:textmaps}). We create a continuous, dense representation to approximate the likelihood with a uniform distribution within an axis-aligned bounding box enclosing all sampled locations where the detection rate is above threshold $\tau$ for this textual cue. 
When collecting posed images is not possible, one can manually annotate these bounding boxes. 


\subsection{MCL Integration}
Our approach extends Monte Carlo localization~\cite{dellaert1999icra}. We provide different sensor models for integrating textual and semantic cues. 

When a textual cue is detected, we inject particles into the corresponding area of the map (\figref{fig:textmaps}), based on the text likelihood maps. 
This technique also allows delocalized robots to get new hypotheses based on the detection, enriching their particle filters with the proposed poses, and is sometimes referred to in literature as reciprocal sampling~\cite{prorok2011iros}.

For 2D object detection predictions, we propose a sensor model based on cosine distance between the detected heading and the semantically-annotated map~\cite{zimmerman2023ral}. 
We also provided a sensor model for 3D bounding box prediction~\cite{zimmerman2023iros}, which considers the IoU between a 3D bounding box prediction and the metric-semantic, accounting for the typical detection noise for that specific semantic class. 
We use the detected semantic objects to infer the room category and utilize this hierarchical semantic information to initialize the particles in the filter in the corresponding rooms.




\vspace{-5pt}

\section{Experimental Evaluation} \label{sec:exp}

In the course of our research, we performed various experiments on datasets recorded for several months, capturing different scenarios such as dynamic obstacles, opening and closing of doors and rearranging furniture.

The outcome of the semantic stability analysis~\cite{zimmerman2023ral}, suggested that it is not necessary to carry this analysis for each type of environment, as the classes that were singled out through rigorous data collection and statistical analysis perfectly matched what we expected to see by using common sense. Additionally, throughout our research, we discovered that the classes chosen for one instance of a specific environment (i.e. "office") transferred successfully to other instances. We used the same classes for deployments in ETH Zurich and ABB facility in Sweden, as well as our later work on semantic localization on resource-constrained platforms~\cite{zimmerman2024icra}, where we ported our human-inspired localization to an ultra-low-power compute platform with 1.5~MB of memory. 

In our localization experiments, the semantically-guided localization outperformed the purely-geometric MCL on all sequences.
Despite the imprecise, hand-annotated maps, we were able to robustly localize long-term, even when objects differed from their actual size by 62.5\%, or up to 1.25~$m$. Similarly, integrating textual cues proved to improve localization stability when localizing in environments with high geometric symmetry and lack of semantic features. Further details on the experiments, qualitative and quantitative results, ablation studies, and insights can be found in the main papers~\cite{zimmerman2023ral, zimmerman2024icra, zimmerman2023iros, zimmerman2022iros}.

\bibliographystyle{plain_abbrv}

\bibliography{glorified,new}

\begin{thebibliography}{10}

\bibitem{bylow2013rss}
E.~Bylow, J.~Sturm, C.~Kerl, F.~Kahl, and D.~Cremers.
\newblock {Real-Time Camera Tracking and 3D Reconstruction Using Signed
  Distance Functions.}
\newblock In {\em Proc.~of Robotics: Science and Systems (RSS)}, volume~2,
  2013.

\bibitem{cadena2016tro}
C.~Cadena, L.~Carlone, H.~Carrillo, Y.~Latif, D.~Scaramuzza, J.~Neira, I.~Reid,
  and J.~Leonard.
\newblock {Past, Present, and Future of Simultaneous Localization And Mapping:
  Towards the Robust-Perception Age}.
\newblock {\em IEEE Trans.~on Robotics (TRO)}, 32:1309--1332, 2016.

\bibitem{cui2021iros}
L.~Cui, C.~Rong, J.~Huang, A.~Rosendo, and L.~Kneip.
\newblock {Monte-Carlo Localization in Underground Parking Lots Using Parking
  Slot Numbers}.
\newblock In {\em Proc.~of the IEEE/RSJ Intl.~Conf.~on Intelligent Robots and
  Systems (IROS)}, 2021.

\bibitem{dellaert1999icra}
F.~Dellaert, D.~Fox, W.~Burgard, and S.~Thrun.
\newblock Monte carlo localization for mobile robots.
\newblock In {\em Proc.~of the IEEE Intl.~Conf.~on Robotics \& Automation
  (ICRA)}, 1999.

\bibitem{grisetti2007tro}
G.~Grisetti, C.~Stachniss, and W.~Burgard.
\newblock {Improved Techniques for Grid Mapping with Rao-Blackwellized Particle
  Filters}.
\newblock {\em IEEE Trans.~on Robotics (TRO)}, 23(1):34--46, 2007.

\bibitem{li2020iros}
L.~Li, B.~Yang, M.~Liang, W.~Zeng, and M.~Ren.
\newblock {End-to-end Contextual Perception and Prediction with Interaction
  Transformer}.
\newblock {\em Proc.~of the IEEE/RSJ Intl.~Conf.~on Intelligent Robots and
  Systems (IROS)}, 2020.

\bibitem{liao2020aaai}
M.~Liao, Z.~Wan, C.~Yao, K.~Chen, and X.~Bai.
\newblock Real-time scene text detection with differentiable binarization.
\newblock In {\em Proceedings of the AAAI conference on artificial
  intelligence}, volume~34, pages 11474--11481, 2020.

\bibitem{loo2024arxiv}
J.~Loo and D.~Hsu.
\newblock {Scene Action Maps: Behavioural Maps for Navigation without Metric
  Information}.
\newblock {\em arXiv preprint arXiv:2405.07948}, 2024.

\bibitem{mendez2018icra}
O.~Mendez, S.~Hadfield, N.~Pugeault, and R.~Bowden.
\newblock {Sedar-semantic detection and ranging: Humans can localise without
  lidar, can robots?}
\newblock In {\em Proc.~of the IEEE Intl.~Conf.~on Robotics \& Automation
  (ICRA)}, 2018.

\bibitem{moravec1989sdsr}
H.P. Moravec.
\newblock {Sensor Fusion in Certainty Grids for Mobile Robots}.
\newblock In {\em Sensor Devices and Systems for Robotics (SDSR)}, 1989.

\bibitem{newcombe2011ismar}
R.A. Newcombe, S.~Izadi, O.~Hilliges, D.~Molyneaux, D.~Kim, A.J. Davison,
  P.~Kohli, J.~Shotton, S.~Hodges, and A.~Fitzgibbon.
\newblock {KinectFusion: Real-Time Dense Surface Mapping and Tracking}.
\newblock In {\em Proc.~of the Intl.~Symposium~on Mixed and Augmented Reality
  (ISMAR)}, pages 127--136, 2011.

\bibitem{prorok2011iros}
A.~Prorok and A.~Martinoli.
\newblock {A reciprocal sampling algorithm for lightweight distributed
  multi-robot localization}.
\newblock In {\em Proc.~of the IEEE/RSJ Intl.~Conf.~on Intelligent Robots and
  Systems (IROS)}, 2011.

\bibitem{radwan2016icra}
N.~Radwan, G.~Tipaldi, L.~Spinello, and W.~Burgard.
\newblock {Do You See the Bakery? Leveraging Geo-Referenced Texts for Global
  Localization in Public Maps}.
\newblock In {\em Proc.~of the IEEE Intl.~Conf.~on Robotics \& Automation
  (ICRA)}, 2016.

\bibitem{rosinol2020icra}
A.~Rosinol, M.~Abate, Y.~Chang, and L.~Carlone.
\newblock {Kimera: an open-source library for real-time metric-semantic
  localization and mapping}.
\newblock In {\em Proc.~of the IEEE Intl.~Conf.~on Robotics \& Automation
  (ICRA)}, 2020.

\bibitem{shi2016tpami}
B.~Shi, X.~Bai, and C.~Yao.
\newblock An end-to-end trainable neural network for image-based sequence
  recognition and its application to scene text recognition.
\newblock {\em IEEE transactions on pattern analysis and machine intelligence},
  39(11):2298--2304, 2016.

\bibitem{tao2024ral}
Y.~Tao, X.~Liu, I.~Spasojevic, S.~Agarwal, and V.~Kumar.
\newblock {3D Active Metric-Semantic SLAM}.
\newblock {\em IEEE Robotics and Automation Letters (RA-L)}, 9(3):2989--2996,
  2024.

\bibitem{thrun2005probrobbook}
S.~Thrun, W.~Burgard, and D.~Fox.
\newblock {\em {Probabilistic Robotics}}.
\newblock MIT Press, 2005.

\bibitem{wada2020cvpr}
K.~Wada, E.~Sucar, S.~James, D.~Lenton, and A.J. Davison.
\newblock {{MoreFusion}: Multi-object Reasoning for {6D} Pose Estimation from
  Volumetric Fusion}.
\newblock In {\em Proc.~of the IEEE/CVF Conf.~on Computer Vision and Pattern
  Recognition (CVPR)}, 2020.

\bibitem{yi2009iros}
C.~Yi, I.H. Suh, G.H. Lim, and B.U. Choi.
\newblock {Bayesian robot localization using spatial object contexts}.
\newblock In {\em Proc.~of the IEEE/RSJ Intl.~Conf.~on Intelligent Robots and
  Systems (IROS)}, 2009.

\bibitem{zafari2019cst}
F.~Zafari, A.~Gkelias, and K.K. Leung.
\newblock {A Survey of Indoor Localization Systems and Technologies}.
\newblock {\em IEEE Communications Surveys Tutorials (CST)}, 21(3):2568--2599,
  2019.

\bibitem{zimmerman2023ral}
N.~Zimmerman, T.~Guadagnino, X.~Chen, J.~Behley, and C.~Stachniss.
\newblock {Long-Term Localization Using Semantic Cues in Floor Plan Maps}.
\newblock {\em IEEE Robotics and Automation Letters (RA-L)}, 8(1):176--183,
  2023.

\bibitem{zimmerman2024icra}
N.~Zimmerman, H.~M{\"u}ller, M.~Magno, and L.~Benini.
\newblock {Fully Onboard Low-Power Localization with Semantic Sensor Fusion on
  a Nano-UAV using Floor Plans}.
\newblock In {\em Proc.~of the IEEE Intl.~Conf.~on Robotics \& Automation
  (ICRA)}. IEEE, 2024.

\bibitem{zimmerman2023iros}
N.~Zimmerman, M.~Sodano, E.~Marks, J.~Behley, and C.~Stachniss.
\newblock {Constructing Metric-Semantic Maps Using Floor Plan Priors for
  Long-Term Indoor Localization}.
\newblock In {\em Proc.~of the IEEE/RSJ Intl.~Conf.~on Intelligent Robots and
  Systems (IROS)}, 2023.

\bibitem{zimmerman2022iros}
N.~Zimmerman, L.~Wiesmann, T.~Guadagnino, T.~Läbe, J.~Behley, and
  C.~Stachniss.
\newblock {Robust Onboard Localization in Changing Environments Exploiting Text
  Spotting}.
\newblock {\em Proc.~of the IEEE/RSJ Intl.~Conf.~on Intelligent Robots and
  Systems (IROS)}, 2022.

\end{thebibliography}

\end{document}